\documentclass{article}

\usepackage{graphicx}
\usepackage{epstopdf}

\usepackage{epsfig} 
\usepackage{amsmath} 
\usepackage{tikz}
\usepackage{textcomp}
\usetikzlibrary{shapes,arrows}
\usetikzlibrary{positioning}

\usepackage[round]{natbib}
\usepackage{mavlab}

\title{Cascaded incremental nonlinear dynamic inversion for MAV disturbance rejection}
\author{Ewoud J.J. Smeur,
Guido C.H.E. de Croon and
Qiping Chu
\thanks{The authors are with the Department
of Aerospace Engineering, Delft University of Technology,
2629HS Delft, The Netherlands. E-mail: e.j.j.smeur@tudelft.nl.}
}

\begin{document}
\maketitle


\begin{abstract}
This paper presents the cascaded integration of Incremental Nonlinear Dynamic Inversion (INDI) for attitude control and INDI for position control of micro air vehicles.
Significant improvements over a traditional Proportional Integral Derivative (PID) controller are demonstrated in an experiment where the quadrotor flies in and out of a 10 m/s windtunnel exhaust.
The control method does not rely on frequent position updates, as is demonstrated in an outside experiment using a standard GPS module.
Finally, we investigate the effect of using a linearization to calculate thrust vector increments, compared to a nonlinear calculation.
\end{abstract}


\tikzset{%
	block/.style    = {draw, thick, rectangle, minimum height = 3em,
	minimum width = 3em},
  sum/.style      = {draw, circle, node distance = 2cm, minimum size = 0.2 cm, text width = 0.3 cm}, 
  gain/.style     = {draw, isosceles triangle},
  input/.style    = {coordinate}, 
  output/.style   = {coordinate}, 
  mux/.style      = {draw, rectangle, fill=black, minimum height =0.5cm, minimum width =0.1cm,inner sep=0}
}
\newcommand{\suma}{$+$}
\newcommand{\minusa}{$^+\! _-$}
\newcommand{\inte}{$\displaystyle \int$}
\newcommand{\derv}{\huge$\frac{d}{dt}$}

\newcommand{\bs}{\boldsymbol}
\newcommand{\pd}[2]{\frac{\partial#1}{\partial#2}}
\newcommand{\si}{\mathrm{s}}
\newcommand{\co}{\mathrm{c}}

\section{Introduction} \label{sec:intro}

Micro Aerial Vehicles (MAV) have the potential to perform many useful tasks, such as search and rescue \citep{ryan}, package delivery \citep{andreapackage}, aerial imaging \citep{imaging}, etc.
In many of these applications, usage of autonomous MAVs can potentially result in significant cost reduction as compared to current practice.
But in order to perform these tasks in an outdoor environment, the vehicles need to be able to control their position under the influence of wind gusts.
This is especially true when flying close to obstacles, as a position error due to a wind gust might result in a collision.

Outdoor MAV missions can encounter significant gusts due to atmospheric turbulence \citep{alexis}.
Moreover, \cite{operations_urban} showed that even a uniform wind can create a very non-uniform wind field in an urban environment.
Computational fluid dynamics calculations showed that with a free stream velocity of 4.6 m/s, flow velocities ranging from 0 to 7.6 m/s are found around buildings.
An MAV flying amidst these buildings can be expected to be subject to up to 7.6 m/s gusts.
For most vehicles this can be considered a strong disturbance, although the effect depends on the weight and surface area of the vehicle.
If such an MAV were to enter a building through an open window in case of a search and rescue mission, it would also experience a sudden change in wind speed.
This scenario is especially challenging due to the confined space of a typical room.
And even indoors, an MAV can be subject to aerodynamic disturbances, for instance caused by its own propeller backwash near walls \citep{shen}.

Clearly, a controller that is able to counteract wind gust disturbances would be of great value.
Currently widespread position control methods such as Proportional Integral Derivative control (PID), which are used even for aggressive control \citep{mellinger}, do not perform well under the influence of gusts.
PID gust rejection properties scale with magnitude of the gains, which is often limited by the GPS update frequency in outdoor scenarios.
Moreover, the integrator term is generally slow in compensating persistent wind disturbances.

To cope with wind gusts, a solution could be to use onboard Pitot tubes to measure the relative velocity of the MAV with respect to the wind.
The difference with the ground speed measured by a GPS module can provide an estimate of the local wind \citep{sidney}.
As disturbances may come from all directions, a minimum of six Pitot tubes would be necessary (two for each axis, as a Pitot tube cannot measure negative airspeed).
Alternatively, \cite{mohamed} used multiple multi-probe sensors to obtain flow pitch angle and velocity.
Adding such an amount of extra sensors will increase the system complexity and cost.
Furthermore, airspeed sensors are typically not reliable at low airspeeds.

Instead of using sensors, one could use a model of the MAV to estimate the wind velocity \citep{escareno}.
\cite{waslander} used an extensive aerodynamic model to estimate wind velocities, with good results in simulation.
The downside of this approach is that it requires a lot of parameters, which might even require windtunnel tests as is done by \cite{schiano} and \cite{tomic}.
If the model does not represent reality well enough due to modeling errors or airframe changes, the gust disturbance rejection performance will degrade.

In this paper, a gust resistant controller is introduced through generalization of Incremental Nonlinear Dynamic Inversion (INDI) to the outer loop control.
The idea is that both disturbances as well as control forces are measured by the accelerometer.
This means that a desired acceleration can be achieved by incrementing the previous control input based on the difference between desired and measured acceleration.
It is shown how to deal with filtering of noisy acceleration measurements, and how this integrates with the INDI attitude controller developed previously \citep{smeur}.
It is also demonstrated that the disturbance rejection capabilities of the INDI inner loop extend to the outer loop control.

\begin{figure}[!t]
	\centering
	\includegraphics[width=\columnwidth]{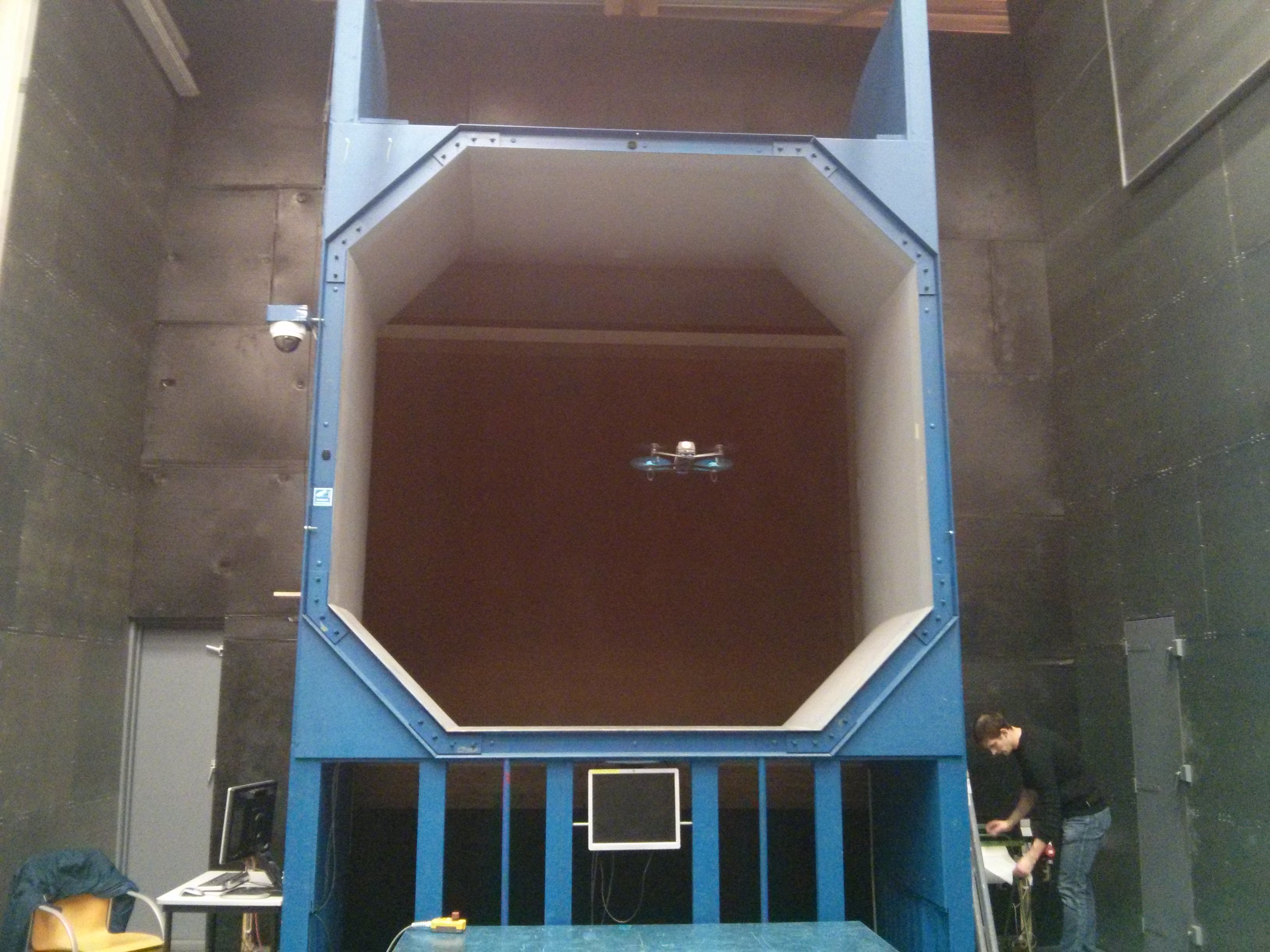}
	\caption{The quadrotor in front of the windtunnel during one of the experiments.}
	\label{fig:experimentfoto}
\end{figure}

The controller is implemented on a Parrot Bebop quadrotor running the Paparazzi open source autopilot software\footnote{The INDI control method is incorporated in Paparazzi, allowing others to easily experiment with it.} \citep{gautierpaparazzi,remes}.
Windtunnel experiments show that the quadrotor can enter and leave the 10 m/s windtunnel flow with only 21 cm maximum position deviation on average.
A controller that uses a gain on the integrated error instead of the incremental controller, suffered 151 cm maximum position deviation upon entering and leaving the windtunnel on average.
To the best of the author's knowledge, this is the first time that a quadrotor is repeatedly flying in and out of a 10 m/s flow as part of a controller's disturbance rejection evaluation.
A picture of the experiment is shown in Figure \ref{fig:experimentfoto}.

This paper is an extension to the work presented at the Intelligent Robots and Systems conference \citep{smeurgust}.
Differences include: (1) the use of a large open jet windtunnel as a more accurate and more powerful disturbance than the fan used previously.
(2) Incorporation of the propeller thrust curve to calculate the total thrust of the drone.
(3) An outdoors experiment based on GPS positioning, to demonstrate the performance in a realistic scenario.
(4) The addition of a nonlinear method to calculate thrust vector increments.

\subsection{Related work}
\cite{hoffmann} developed an altitude controller that utilizes the vertical acceleration measurement.
However, they fed the acceleration back, multiplied with a gain, without utilizing the physical relation between thrust and acceleration.
In a different paper, they state that their PID position control implementation has little ability to reject disturbances from wind and translational velocity effects \citep{hoffmann2}.
A vertical controller using the INDI principle was developed for a traditional helicopter in simulation by \cite{simplicio}.
Only very limited sensor noise was taken into account, which did not require any filtering.
Also, in both of these papers, by separating the vertical axis from the lateral axes, coupling can be expected.
In this paper it is shown that by inverting the control effectiveness for all axes, accelerations in each of these axes can be controlled.

\cite{wang} applied an acceleration feedback dynamic inversion approach to all axes of a quadrotor, and demonstrated accurate trajectory tracking capabilities.
They mentioned robustness against disturbances, but did not analyze or demonstrate the controller response against disturbances.
Also, the effects of accelerometer noise or filtering are not discussed.
Additional differences with the work presented here are that with INDI there is no need for a reference model or command filtering, and that the approach of Wang is not incremental.
This means that if a certain control input does not completely resolve a measured acceleration error (due to input modeling errors or uncertainties), the error will persist.
In an incremental scheme the input can be incremented again to resolve angular acceleration errors.


\section{Incremental nonlinear dynamic inversion for attitude control} \label{sec:indi_att}

\begin{figure}[!thb]
	\centering
	\includegraphics[width=\columnwidth]{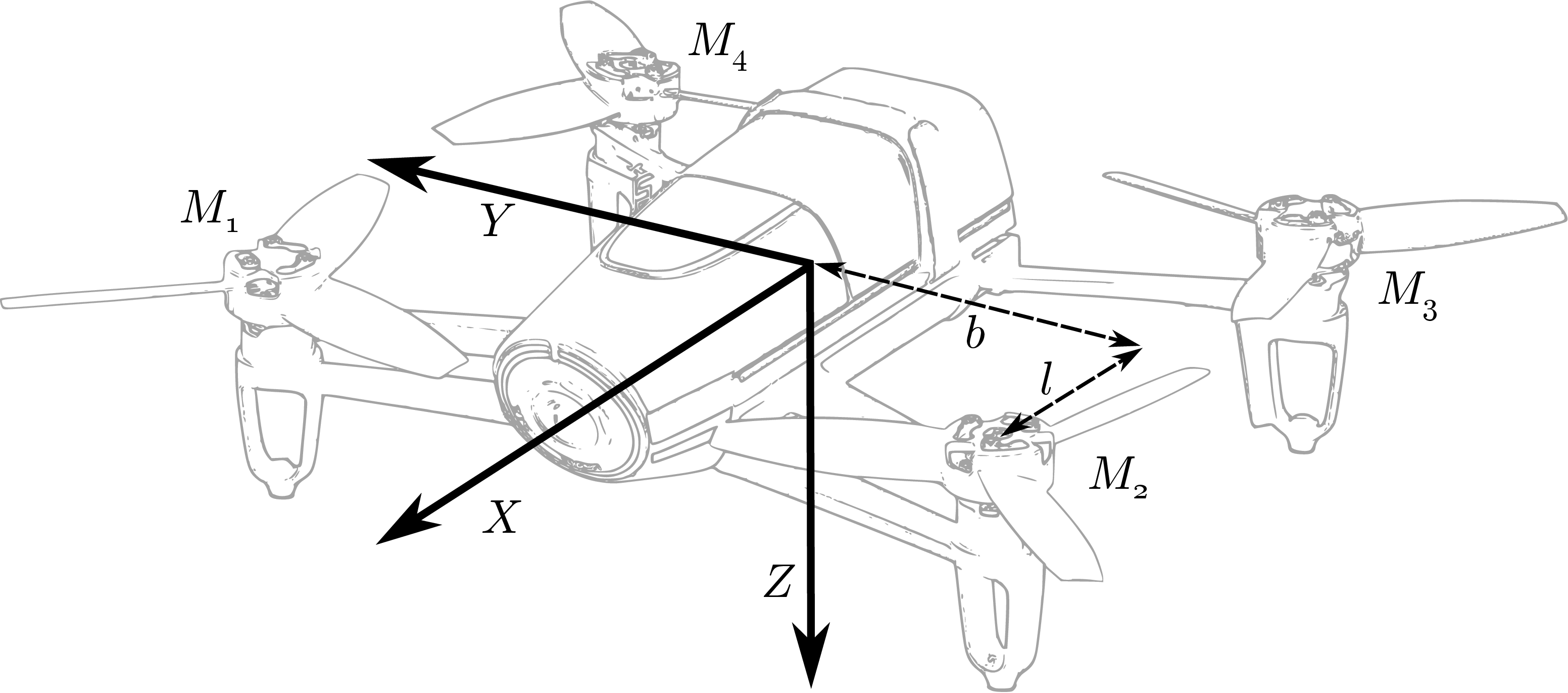}
	\caption{The Bebop Quadcopter used in the experiments with body axis definitions.}
	\label{fig:bebop}
\end{figure}

An extended analysis of INDI for attitude control of MAVs is provided in previous work \citep{smeur}.
For completeness, an overview of the developed attitude controller, along with some new additions, is presented in this section.
Consider the quadrotor shown in Figure \ref{fig:bebop}.
The distance from the center of gravity to each of the rotors along the $X$ axis is given by $l$ and along the $Y$ axis by $b$.

It is defined that $\bs{\Omega}$ is the angular rate vector of the vehicle and the angular rates of the propellers around the body Z axis are described with the vector $\bs{\omega}$.
The thrust provided by all four rotors is denoted by $T$.
Consider Eq. \ref{eq:quadcontrol}, which gives an expression for the angular acceleration and the thrust:

\begin{equation}
	\left[ \begin{array}{c}\dot{\bs{\Omega}} \\ T\end{array} \right]=
		\bs{F}(\bs\Omega,\bs{v}) + \bs{G}(\bs{\omega},\bs{\dot{\omega}})
	\label{eq:quadcontrol}
\end{equation}

Here, $\bs{F}(\bs\Omega,\bs{v})$ is the function that describes the vehicle moments as a function of the angular rates and velocity.
$\bs{G}(\bs{\omega},\bs{\dot{\omega}})$ is the function that maps the input and the derivative of the input to the angular acceleration and thrust.
Note that the thrust force only depends on the rotational rate of the rotors, and so, the fourth row of the $\bs{F}(\bs\Omega,\bs{v})$ matrix is zero.
Now a first order Taylor expansion can be applied:

\begin{equation}
	\begin{array}{rl}
		\left[ \begin{array}{c}\dot{\bs{\Omega}} \\ T\end{array} \right]=&
			\bs{F}(\bs{\Omega}_0,\bs{v}_0) + \bs{G}(\bs{\omega}_0,\bs{\dot{\omega}}_0) \\
		& + \pd{}{\bs{\Omega}} (\bs{F}(\bs{\Omega},\bs{v}_0))|_{\bs{\Omega} = \bs{\Omega}_0}(\bs{\Omega} - \bs{\Omega}_0) \\
		& + \pd{}{\bs{v}} (\bs{F}(\bs{\Omega}_0,\bs{v}))|_{\bs{v} = \bs{v}_0}(\bs{v} - \bs{v}_0) \\
		& + \pd{}{\bs{\omega}} (\bs{G}(\bs{\omega},\bs{\dot{\omega}}_0))|_{\bs{\omega} = \bs{\omega}_0}(\bs{\omega} - \bs{\omega}_0) \\
		& + \pd{}{\bs{\dot{\omega}}}(\bs{G}(\bs{\omega}_0,\bs{\dot{\omega}}))|_{\bs{\dot{\omega}} = \bs{\dot{\omega}}_0}(\bs{\dot{\omega}} - \bs{\dot{\omega}}_0)
	\end{array}
	\label{eq:taylor_att}
\end{equation}

First it can be recognized that the first two terms give the current angular acceleration and thrust:
$\bs{F}(\bs{\Omega}_0,\bs{v}_0) + \bs{G}(\bs{\omega}_0,\bs{\dot{\omega}}_0) =
\begin{array}{lr}[\dot{\bs{\Omega}}^T_0 & T_0]^T\end{array}$.
The next two terms predict the change in moment due to changes airspeed and rotational rate.
For these terms, both a detailed model and an airspeed estimate are needed.
In absence of this information, these terms are omitted to simplify the equation and as such the terms are treated as a disturbance.
This may lead to small errors in the angular acceleration prediction.
However, moments due to these terms eventually show up in the measured angular acceleration and are taken into account this way.
Finally, it is assumed that over the operational domain, the partial derivatives of $\bs{G}(\bs{\omega},\dot{\bs{\omega}})$ do not change.
Therefore, they can be approximated by the static matrices $\bs{G}_1$ and $\bs{G}_2$.
These control effectiveness matrices are (4$\times$4), because they contain the effectiveness of each of the four rotors on each of the axes roll, pitch, yaw and thrust.

\begin{equation}
	\left[ \begin{array}{c}\dot{\bs{\Omega}} - \dot{\bs{\Omega}}_0 \\ T - T_0\end{array} \right]=
		\bs{G}_1(\bs{\omega} - \bs{\omega}_0) + T_s\bs{G}_2(\bs{\dot{\omega}} - \bs{\dot{\omega}}_0)
	\label{eq:partialsgone}
\end{equation}

Here, the sample time $T_s$ is factored out of $\bs{G}_2$ to simplify future calculations.
The angular acceleration $\bs{\dot{\Omega}}_0$ can be determined, by deriving it from the gyroscope using finite difference.
This signal is often very noisy, because the rotating propellers lead to vibrations in the airframe.
From \cite{bacon}, the use of a second order filter is adopted, given by:

\begin{equation}
	H(s) = \frac{\omega_n^2}{s^2 + 2\zeta\omega_n s + \omega_n^2}
	\label{eq:sec_filter}
\end{equation}

This filter also introduces a delay.
To be able to apply the Taylor expansion, terms with index 0 should be from the same moment in time.
This is why all these terms should be filtered with the same filter, such that they are equally delayed.
These terms are given the subscript $f$ in Eq. \ref{eq:filtered}.

\begin{equation}
	\left[ \begin{array}{c}\dot{\bs{\Omega}} - \dot{\bs{\Omega}}_f \\ T - T_f\end{array} \right]=
		\bs{G}_1(\bs{\omega} - \bs{\omega}_f) + T_s\bs{G}_2(\bs{\dot{\omega}} - \bs{\dot{\omega}}_f)
	\label{eq:filtered}
\end{equation}

By approximating $\dot{\bs{\omega}}$ and $\dot{\bs{\omega}}_f$ with finite difference, using the lag operator $L$ as $\dot{\bs{\omega}} = (\bs{\omega}(k) - \bs{\omega}(k-1))/T_s = (\bs{\omega} - L \bs{\omega})/T_s$
and rearranging the terms, this equation becomes:

\begin{equation}
	\arraycolsep=1.4pt
	\left[ \begin{array}{c}\dot{\bs{\Omega}} - \dot{\bs{\Omega}}_f \\ T - T_f\end{array} \right]=
		(\bs{G}_1 + \bs{G}_2 ) (\bs{\omega} - \bs{\omega}_f)
		- \bs{G}_2 L (\bs{\omega} - \bs{\omega}_{f})
	\label{eq:finaltaylor}
\end{equation}

This equation can now be inverted, to yield Eq. \ref{eq:indi_att}.

\begin{equation}
	\arraycolsep=1.4pt
	\begin{array}{ll}
		\bs{\omega}_c  = \bs{\omega}_f +& (\bs{G}_1 + \bs{G}_2)^{-1} \\  & \cdot \left(
		\left[ \begin{array}{c} \bs{\nu}_{\dot{\bs{\Omega}}} - \bs{\dot{\Omega}}_f \\ \widetilde{T} \end{array} \right]
		+ \bs{G}_2 L (\bs{\omega}_c - \bs{\omega}_{f})\right)
	\end{array}
	\label{eq:indi_att}
\end{equation}

Here, $\bs{\nu}_{\dot{\bs{\Omega}}}$ is the virtual control, which is the desired angular acceleration that has now become an input.
In the next section it will be shown that the thrust increment $T-T_f$ is calculated in the outer loop, and is therefore denoted by $\widetilde{T}$.
The subscript $c$ is added to $\bs{\omega}$ to indicate that this is the command sent to the motors.

The control diagram is shown in Figure \ref{fig:indi_inner}.
The input to this diagram is the angular acceleration virtual control and the output is the angular acceleration of the vehicle.
The angular acceleration error, and the thrust increment go into the inversion of the control effectiveness.
The increment in motor command is added to the feedback from the motors, which is filtered with the same filter as the angular acceleration.
The actuator dynamics are denoted by $A(z)$.

\begin{figure*}
	\centering
	\begin{tikzpicture}[auto, node distance=1cm, >=triangle 45]
		\draw
		node [input, name=input1] {}
		node [sum, right=0.8cm of input1] (suma1) {\minusa}
		node [input, right= 0.9cm of suma1] (dummy1) {}
		node [input, above= 0.30cm of dummy1] (dummy2) {}
		node [mux, below = 0.05cm of dummy2] (mux1) {}
		node [mux, above = -0.1cm of dummy2] (spacefiller) {}
		node [mux, above = 0.05cm of dummy2] (mux2) {}
		node [sum, right=0.6cm of dummy2] (sumc) {\suma}
		node [block, right=0.6cm of sumc] (k) { $(\bs{G}_1 + \bs{G}_2)^{-1}$}
		node [input, right= 0.4cm of k] (space4) {}
		node [sum, right=0.5cm of space4] (sumb1){\suma}
		node [input, right=0.7cm of sumb1] (space1) {}
		node [block, below=0.82cm of space1] (hs) {$\bs{H}(z)$}
		node [block, right=1.4cm of space1] (AC) {$\bs{A}(z)$}
		node [block, below right=0.82cm and 1.4cm of space1] (integratorac) {$\frac{1}{z}$}
		node [block, right=0.8cm of AC] (mav) {MAV}
		node [input, above=0.65cm of suma1] (specific_force_in) {}
		node [input, right=0.8cm of mav] (spaceend) {}
		node [block, below=1.0cm of spaceend] (integrator) {$\frac{T_s z}{z - 1} \frac{1}{z}$}
		node [input, below= 0.82cm of k] (space3) {}
		node [gain, rotate=180, left= 1.9cm of space3] (l) {\rotatebox{180}{$\bs{G}_2$}}
		node [block, left=-1.1cm of space3] (delay) {$\frac{1}{z}$}
		node [block, below=0.4cm of hs] (hs2) {$\bs{H}(z)$}
		node [block, left=2.55cm of hs2] (difference) {$\frac{z-1}{T_sz}$}
		node [output, right=1.4cm of spaceend] (out) {};
		\draw[->](input1) -- node {$\boldsymbol{\nu}_{\dot{\bs{\Omega}}}$}(suma1);
		\draw[->](suma1) -- node {$\dot{\boldsymbol{\Omega}}_\mathrm{err}$} (mux1);
		\draw[->](dummy2) -- node {} (sumc);
		\draw[->](sumc) -- node {} (k);
		\draw[->](k) -- node {$\widetilde{\boldsymbol{\omega}}$} (sumb1);
		\draw[->](sumb1) -- node {$\bs{\omega}_c$} (AC);
		\draw[->](hs) -| node{$\boldsymbol{\omega}_f$} (sumb1);
		\draw[->]([xshift=0.4cm]AC.east) |- node{$\bs{\omega}$} (integratorac.east);
		\draw[->](AC) -- node{} (mav);
		\draw[->]([xshift=0.5cm]specific_force_in) |- node[xshift=0.3cm]{$\widetilde{T}$} (mux2);
		\draw[->](integratorac) -- node{} (hs);
		\draw[->](mav) -| node{} (integrator);
		\draw[->](integrator) |- node{$\boldsymbol{\Omega}_0$} (hs2);
		\draw[->](hs2) -- node{$\bs{\Omega}_f$} (difference);
		\draw[->](difference) -| node{$\boldsymbol{\dot{\Omega}}_f$} (suma1);
		\draw[->](spaceend) -- node{$\bs{\dot{\Omega}}$} (out);
		\draw[->](space4) |- node{} (delay);
		\draw[->](delay) -- node{} (l);
		\draw[->](l) -| node{} (sumc);

		\draw
		;
	\end{tikzpicture}
	\caption{The inner INDI control structure.}
	\label{fig:indi_inner}
\end{figure*}

\subsection{Attitude control}

The control law of Eq. \ref{eq:indi_att} describes how to track angular accelerations.
A PD controller can be used to provide the angular acceleration that will steer the vehicle towards a desired attitude.
For the attitude feedback, a quaternion representation is used, as described by \cite{fresk}.
The calculation of the error quaternion $\bs{q}_\text{err}$ in terms of the reference $\bs{q}_\text{ref}$ and the state quaternion $\bs{q}_\text{s}$ is then:
\begin{equation}
	\bs{q}_\text{err} = \bs{q}_\text{ref} \otimes \bs{q}^*_\text{s}
	\label{eq:quaterr}
\end{equation}
where $\otimes$ denotes the Hamilton product and $^*$ denotes conjugation.
The reference angular rate is found using the vector part of the quaternion:
\begin{equation}
\bs{\Omega}_\text{ref} = K_\eta \left[ \begin{array}{ccc} q_1^\text{err} & q_2^\text{err} & q_3^\text{err} \end{array} \right]^T
	\label{eq:omegaerr}
\end{equation}
The angular acceleration reference is then calculated from the rotational rate error, using a second gain $K_\Omega$.

In order to find a theoretical basis for what the values of these gains should be, linear time-invariant systems theory will be used.
That means that the attitude feedback needs to be simplified a bit, as is shown in Figure \ref{fig:attitude}.
In this figure small angles are assumed, in order to allow simple integration of the rates to obtain the attitude $\bs{\eta}$ in Euler angles.

\begin{figure*}[h!]
\centering
\begin{tikzpicture}[auto, thick, node distance=2cm, >=triangle 45]
\draw
  node [input, name=input1] {}
  node [sum, right=1.0cm of input1] (sumb1){\minusa}
	node [gain, right=1.0cm of sumb1] (keta){$\frac{1}{2}K_\eta$}
  node [sum, right=0.8cm of keta] (sumb2){\minusa}
  node [gain, right=1.0cm of sumb2] (komega){$K_\Omega$}
  node [block, right=1.0cm of komega] (AC) {$\bs{A}(z)$}
  node [block, right=1.0cm of AC] (integrator1) {$\frac{T_s z}{z - 1}$}
  node [block, right=1.0cm of integrator1] (integrator2) {$\frac{T_s z}{z - 1}$}
  node [output, right=1.5cm of integrator2] (out) {};
  \draw[->](input1) -- node {$\bs{\eta}_{\mathrm{ref}}$}(sumb1);
  \draw[->](sumb1) -- node {} (keta);
  \draw[->](keta) -- node {$\bs{\Omega}_{\mathrm{ref}}$} (sumb2);
  \draw[->](sumb2) -- node {} (komega);
  \draw[->](komega) -- node {} (AC);
  \draw[->](AC) -- node {$\dot{\bs{\Omega}}$} (integrator1);
  \draw[->](integrator1) -- node {$\bs{\Omega}$} (integrator2);
  \draw[->]([xshift=0.5cm]integrator1.east) -- ++(0,-1cm) -| node{} (sumb2);
  \draw[->]([xshift=0.5cm]integrator2.east) -- ++(0,-1.7cm) -| node{} (sumb1);
  \draw[->](integrator2) -- node{$\bs{\eta}$} (out);

  \draw [color=gray,thick](9.1,1.2) rectangle (11.0,-0.7);
  \node at (-0.5,1) [above=2mm, right=9.9cm] {\textsc{INDI}};
\draw
;
\end{tikzpicture}
\caption{The design of the attitude controller for small angles, based on the closed loop response of the INDI controller. The feedback of the attitude is simplified for the sake of the analysis, and proper quaternion feedback is used on the real platform.}
\label{fig:attitude}
\end{figure*}

Because the proper quaternion integration is removed in Figure \ref{fig:attitude}, there is a factor $\frac{1}{2}$ with $K_\eta$, since the quaternion derivative is defined as:
\begin{equation}
\dot{\bs{q}} = \frac{1}{2} \bs{q} \otimes \left[ \begin{array}{c} 0 \\ \bs{\Omega} \end{array} \right]
	\label{eq:quat_derivative}
\end{equation}

In previous research \citep{smeur} it was shown that if the assumptions, mentioned in the derivation of the controller, hold true, the transfer function from $\boldsymbol{\nu}_{\dot{\bs{\Omega}}}$ to $\dot{\bs{\Omega}}$ is simply the actuator dynamics $\bs{A}(z)$.
When the actuator dynamics can be modeled, for instance by first order dynamics, the P ($K_\eta$) and D ($K_\Omega$) gains can be determined based on the desired poles and zeros of the system.
For the Bebop, the actuator dynamics are modeled with first order dynamics as shown in Eq. \ref{eq:act_dyn}, with $\alpha = 0.1$ at a sample frequency of 512 Hz.
\begin{equation}
	A(z) = \frac{\alpha}{z - (1-\alpha)}
	\label{eq:act_dyn}
\end{equation}
Then the transfer function of the closed loop system from Figure \ref{fig:attitude} is as follows:
\begin{equation}
	\small
	\begin{array}{l}
		TF_{\eta_{\mathrm{ref}} \rightarrow \eta} =\\
		\frac{{{K_\eta }{K_\Omega }\alpha T_s^2{z^2}}}{{{z^3} + \left( {{K_\Omega }\alpha {T_s} + {K_\eta }{K_\Omega }\alpha T_s^2 + \alpha  - 3} \right){z^2} + \left( {3 - 2\alpha  - {K_\Omega }\alpha {T_s}} \right)z - 1 + \alpha }}
	\end{array}
	\label{eq:transfer}
\end{equation}
The gains can now be selected such that the poles are within the unit circle and the response is fast with little overshoot.
With $K_\Omega = 28.0$ and $K_\eta = 21.4$, there is one pole at 0.964 and two complex poles at $0.965\pm0.0445i$.

\begin{figure}[htb]
	\centering
	\includegraphics[width=\columnwidth]{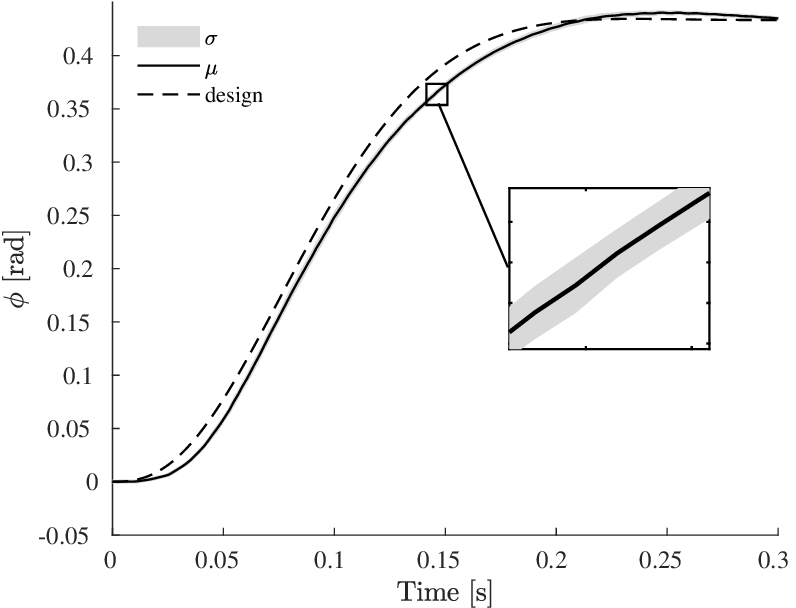}
	\caption{Comparison of the designed response and the actual response of the attitude of the quadrotor. The black line is the average, and the gray area one standard deviation of 25 repetitions.}
	\label{fig:attstep}
\end{figure}

An interesting question is how close the theoretically designed response of the attitude matches the actual attitude response of the quadrotor.
To test this, both the above transfer function and the real quadrotor are subjected to a step input.
For the real quadrotor, the step input is repeated 25 times, and the mean and standard deviation are shown in Figure \ref{fig:attstep}.
The response of the above transfer function is shown in the same figure in red.
The difference between the designed response and the actual response is rather small: the error as a percentage of the step magnitude is maximum 6.4 \% at 0.14 s.
This means that this is a valid way of designing the P and D gains, based solely on the first order actuator dynamics model.

Some may argue that a nonlinear Lyapunov stability proof is more convincing than what is presented here.
Note that, using a Lyapunov proof, the stability of a very similar quaternion feedback law was already proven by \cite{tayebi}.
Many recent papers on the topic of MAV control contain such a proof \citep{lee,wen,mian,yu}, and in each of these papers, an equivalent to the P and D gains exists.
Further, in each of these papers, the claim is made, either explicit or implicit, that these gains can be chosen freely, irrespective of the Lyapunov proof.
One might even get away with the feeling, reading these papers, that these gains have little influence on the final performance.

As anyone who has practical experience with control of drones can attest, these P and D gains are, in reality, not free to choose without consequences for the stability.
There are most certainly non-negative, non-zero values of these gains for which the system is not stable, especially if the actuator dynamics are slow.
This discrepancy exists because in these papers the actuator dynamics are always neglected.
As is shown in this section, the actuator dynamics are crucial for the performance, as well as the stability.
For different actuator dynamics, for example $\alpha = 0.02$, the transfer function of Equation \ref{eq:transfer} would even be unstable with the same gains.

Of course, it is still a good thing to be able to prove Lyapunov stability, even while neglecting actuator dynamics.
It at least indicates that feedback is applied in the right direction, which may be troublesome in the case of feedback of Euler angles.
However, that is not an answer to the complete stability question.
Taking the actuator dynamics into account, for instance using the linear methods employed here, is an indispensable part of the stability and performance analysis.





\subsection{Determining the control effectiveness} \label{sec:ctrl_eff}

The control effectiveness can be derived from detailed knowledge of the system inertia and actuator forces.
However, accurately measuring the vehicle inertia can prove to be difficult, time consuming and not very accurate.
Alternatively, the control effectiveness matrices can be estimated using test flight data.
A single test flight, with both the actuator inputs as well as the gyroscope and accelerometer data logged at high frequency, can provide enough information to determine the values for $\bs{G}_1$ and $\bs{G}_2$.
Though some effort is required to make the vehicle fly stable without knowledge of the control effectiveness, this can be a very fast and efficient method to obtain the effectiveness.
Further, the flight data will represent the system as it actually is, eliminating chances for modeling errors.

When the flight has been recorded, the control effectiveness can simply be calculated by finding the least squares solution of:
\begin{equation}
\left[ \begin{array}{c} \ddot{\bs{\Omega}}_f \\ \dot{T}_f\end{array} \right]=
\left[\begin{array}{cc} \bs{G}_1 & T_s\bs{G}_2 \end{array}\right] \left[\begin{array}{c} \dot{\bs{\omega}}_f \\ \bs{\ddot{\omega}}_f \end{array}\right]
	\label{eq:estimation}
\end{equation}
where the derivatives can be calculated using the method of finite difference.

\section{Incremental nonlinear dynamic inversion applied to linear accelerations} \label{sec:indi}

Now that the attitude of the quadrotor is controlled, it is possible to derive an incremental controller for the linear acceleration of the vehicle.
Two reference frames will be used throughout this derivation; the body frame, as depicted in Figure \ref{fig:bebop}, and the North East Down (NED) frame, which has its origin fixed to a point on the earth.
Vectors in the body frame have a subscript $B$ and vectors in the NED frame have subscript $N$.
The subscripts will only be used to avoid confusion, the position $\bs{\xi}$ and velocity $\dot{\bs{\xi}}$ of the quadrotor will always be in the NED frame.

The position dynamics are given by Newton's second law of motion:

\begin{equation}
	\ddot{\bs{\xi}} = \bs{g} + \frac{1}{m}\bs{F}(\dot{\bs{\xi}},\bs{w}) + \frac{1}{m}\bs{T}_{N}(\bs{\eta},T)
	\label{eq:newton}
\end{equation}

Where $\ddot{\bs{\xi}}$ is the acceleration of the MAV, $\bs{g}$ is the gravity vector and $m$ is the mass.
$\bs{F}$ is the aerodynamic force working on the airframe as a function of the velocity $\dot{\bs{\xi}}$ of the MAV and the wind vector $\bs{w}$.
$\bs{T}_{N}$ is the thrust vector in the NED frame as a function of the attitude $\bs{\eta}= [\phi, \theta, \psi]^T$ and the total thrust produced by the four rotors $T$.

The thrust vector in the NED frame can be obtained by taking the thrust vector in the body frame, given by $\bs{T}_{B} = [0, 0, T]^T$, and rotating it using the rotation matrix $\bs{M}_{NB}(\bs{\eta})$.
Since the thrust vector in the body frame only has a $Z$ component, only the last column of the rotation matrix is relevant.
The thrust vector in the NED frame is therefore given by:

\begin{equation}
	\bs{T}_N(\bs{\eta},T) = \bs{M}_{NB}(\bs{\eta})\bs{T}_B =
	\left[
		\begin{array}{c}
			(\si\phi\si\psi + \co\phi\co\psi\si\theta)T \\
			(\co\phi\si\psi\si\theta - \co\psi\si\phi)T \\
			(\co\phi\co\theta)T
		\end{array}\right]
	\label{eq:tn}
\end{equation}

where the sine and cosine functions are abbreviated by the letters s and c respectively.
Now a first order Taylor expansion can be applied to Eq. \ref{eq:newton}, resulting in Eq. \ref{eq:taylor}:

\begin{equation}
	\begin{array}{ll}
		\ddot{\bs{\xi}} =&
		\bs{g} + \frac{1}{m}\bs{F} (\dot{\bs{\xi}}_0,\bs{w}_0) + \frac{1}{m} \bs{T}_N(\bs{\eta}_0,T_0) \\
		&+ \pd{}{\dot{\bs{\xi}}} \frac{1}{m} \bs{F}(\dot{\bs{\xi}},\bs{w}_0)|_{\dot{\bs{\xi}} = \dot{\bs{\xi}}_0} (\dot{\bs{\xi}} - \dot{\bs{\xi}}_0)\\
		&+ \pd{}{\bs{w}} \frac{1}{m} \bs{F}(\dot{\bs{\xi}}_0,\bs{w})|_{\bs{w} = \bs{w}_0} (\bs{w} - \bs{w}_0)\\
		&+ \pd{}{\phi}  \frac{1}{m} \bs{T}_N(\phi,\theta_0,\psi_0,T_0)|_{ \phi= \phi_0} (\phi - \phi_0)\\
		&+ \pd{}{\theta}  \frac{1}{m} \bs{T}_N(\phi_0,\theta,\psi_0,T_0)|_{ \theta= \theta_0} (\theta - \theta_0)\\
		&+ \pd{}{\psi}  \frac{1}{m} \bs{T}_N(\phi_0,\theta_0,\psi,T_0)|_{ \psi= \psi_0} (\psi - \psi_0)\\
		&+ \pd{}{T}  \frac{1}{m} \bs{T}_N(\phi_0,\theta_0,\psi_0,T)|_{ T= T_0} (T - T_0)
	\end{array}
	\label{eq:taylor}
\end{equation}

The first terms can be simplified to the acceleration at the previous timestep: $\bs{g} + \frac{1}{m} \bs{F} (\dot{\bs{\xi}}_0) + \frac{1}{m} \bs{T}_N(\bs{\eta}_0,T_0) = \ddot{\bs{\xi}}_0$.
This acceleration can be obtained by rotating the specific force measured by the accelerometer in the body axes to the NED frame and adding the gravity vector.
For the next two terms, the partial derivative of $\bs{F}$ with respect to $\dot{\bs{\xi}}$ and $\bs{w}$, there is not a good estimate.
For simplicity of the approach, the choice is made not to employ a model of the aerodynamic drag of the airframe.
Moreover, it is very difficult, if not impossible, to predict how the wind is going to change.
Therefore, the best guess for these terms is zero.
Note that this does not mean that all aerodynamic forces are neglected.
These forces will be measured with $\ddot{\bs{\xi}}_0$.
Finally, it is assumed that changes in $\psi$ will be small, such that this term can be neglected.
Combining this with Eq. \ref{eq:tn} and \ref{eq:taylor} leads to:
\begin{equation}
	\ddot{\bs{\xi}} = \ddot{\bs{\xi}}_0 + \frac{1}{m} G(\bs{\eta}_0,T_0)  (\bs{u} - \bs{u}_0)
	\label{eq:taylorresult}
\end{equation}
where $\bs{u} = \begin{array}{lcr}[\phi & \theta & T]^T
\end{array}$
and
\small
\arraycolsep=3pt 
\begin{equation}
	\label{eq:gtaylor}
	\begin{array}{l}
		\bs{G}(\bs{\eta},T) = \\
		\left[
			\begin{array}{ccc}
				(\co\phi\si\psi - \si\phi\co\psi\si\theta)T & (\co\phi\co\psi\co\theta)T & \si\phi\si\psi + \co\phi\co\psi\si\theta \\
				(-\si\phi\si\psi\si\theta - \co\psi\co\phi)T & (\co\phi\si\psi\co\theta)T & \co\phi\si\psi\si\theta - \co\psi\si\phi \\
				-\co\theta\si\phi T & -\si\theta\co\phi T & \co\phi\co\theta
			\end{array}
		\right]
	\end{array}
\end{equation}
\normalsize

The measured accelerations, necessary to obtain $\ddot{\bs{\xi}}_0$, are typically noisy due to vibrations in the airframe introduced by the spinning propellers.
Therefore, the accelerations need to be filtered.
Like in the previous section, the delay of the filter needs to be accounted for.
This is why also here, all terms with subscript 0 will be filtered with the same filter, and be given a subscript $f$.
Then, if Eq. \ref{eq:taylorresult} is inverted, the INDI control law for linear accelerations is obtained:

\begin{equation}
	\bs{u}_c  = \bs{u}_f + m G^{-1}(\bs{\eta}_0,T_0)  (\bs{\nu}_{\ddot{\xi}} - \ddot{\bs{\xi}}_f)
	\label{eq:indi}
\end{equation}

$\ddot{\bs{\xi}}$ is replaced with the virtual control $\bs{\nu}_{\ddot{\xi}}$ to indicate that this is now an input to the equation (the desired acceleration), and the subscript $c$ is added to $\bs{u}$ to indicate that this is the command that will be sent to the inner loop controller.
Also, the control increment is defined to be $\widetilde{\bs{u}} = \bs{u}_c - \bs{u}_f$, so clearly Eq. \ref{eq:indi} is an incremental control law.

Suppose the inner loop is filtered with filter $f_1$, and the outer loop with filter $f_2$.
The thrust increment required by the inner loop is then $T-T_{f_1}$.
The thrust increment calculated by the outer loop is $T-T_{f_2}$.
It is only possible to pass on the thrust increment from the outer to the inner loop if these filters are equal.
That is why for both loops, the filter described by Eq. \ref{eq:sec_filter} is used, with the same parameters.

\begin{figure*}
	\centering
	\begin{tikzpicture}[auto, node distance=1cm, >=triangle 45]
		\draw
		node [input, name=input1] {}
		node [sum, right=0.8cm of input1] (suma1) {\minusa}
		node [block, right=0.8cm of suma1] (k) { $m\bs{G}(\bs{\eta}_0,T_0)^{-1}$}
		node [input, right= 0.7cm of k] (dummy1) {}
		node [mux, below = 0.cm of dummy1] (mux1) {}
		node [mux, above = -0.cm of dummy1] (mux2) {}
		node [sum, right=1.0cm of mux1] (sumb1){\suma}
		node [input, right=0.7cm of sumb1] (space1) {}
		node [block, below=1.2cm of space1] (hs) {$\bs{H}(z)$}
		node [block, right=3.6cm of dummy1] (inner_loop) {Inner loop}
		node [block, right=1.4cm of hs] (integratorac) {$\frac{1}{z}$}
		node [block, right=0.7cm of inner_loop] (g) {MAV}
		node [input, right=1.4cm of g] (spaceend) {}
		node [block, below=1.0cm of spaceend] (integrator) {$\frac{1}{z}$}
		node [input, below= 0.82cm of k] (space3) {}
		node [block, below=0.3cm of hs] (hs2) {$\bs{H}(z)$}
		node [output, right=1.4cm of spaceend] (out) {};
		\draw[->](input1) -- node {$\boldsymbol{\nu}_{\ddot{\xi}}$}(suma1);
		\draw[->](k) -- node {} (dummy1);
		\draw[->](suma1) -- node {$\ddot{\boldsymbol{\xi}}_\mathrm{err}$} (k);
		\draw[->](mux1) -- node[yshift=-1.3cm] {$\left[\begin{array}{c} \widetilde{\phi} \\ \widetilde{\theta} \end{array} \right]$} (sumb1);
		\draw[->](sumb1) -- node[yshift=-1.1cm] {$\left[\begin{array}{c} \phi_c \\ \theta_c \end{array} \right]$} (sumb1 -| inner_loop.west);
		\draw[->](hs) -| node[yshift=-0.5cm,xshift=0.2cm]{$\left[\begin{array}{c} \phi_f \\ \theta_f \end{array} \right]$} (sumb1);
		\draw[->](g.south) |- node{$\left[\begin{array}{c} \phi \\ \theta \end{array} \right]$} (integratorac.east);
		\draw[->](mux2) -- node{$\widetilde{T}$} (mux2 -| inner_loop.west);
		\draw[->](integratorac) -- node{} (hs);
		\draw[->](inner_loop) -- node{} (g);
		\draw[->](g) -| node{} (integrator);
		\draw[->](integrator) |- node{$\ddot{\bs{\xi}}_0$} (hs2);
		\draw[->](hs2) -| node{$\ddot{\bs{\xi}}_f$} (suma1);
		\draw[->](spaceend) -- node{$\ddot{\bs{\xi}}$} (out);
		\draw
		;
	\end{tikzpicture}
	\caption{The outer INDI control structure.}
	\label{fig:indi_outer}
\end{figure*}
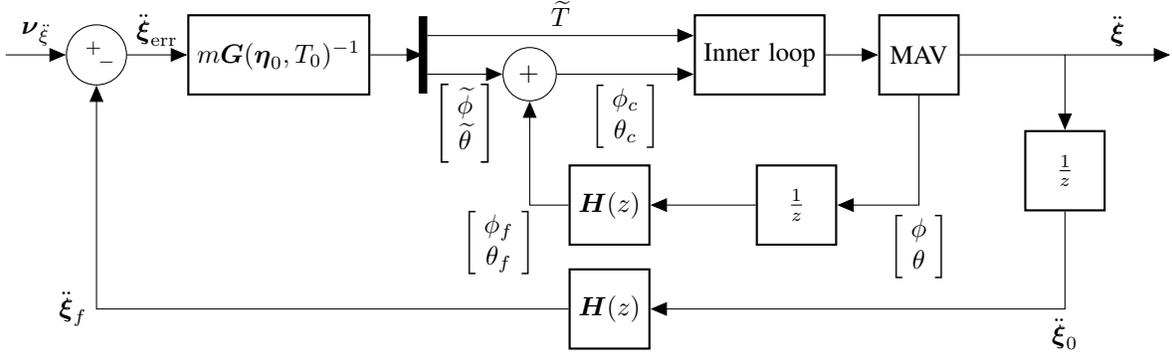

\section{Implementation} \label{sec:implementation}

The implementation of the control law given by Eq. \ref{eq:indi} is shown in Figure \ref{fig:indi_outer}.
The input of this diagram is the virtual control, and the output is the acceleration in NED frame.
The acceleration in NED frame can be obtained from the accelerometer measurements with a simple rotation matrix and the addition of the gravity vector.
Increments in roll, pitch and thrust are obtained from the error in acceleration multiplied with the inverse of the control effectiveness matrix.
The roll and pitch increments are added to the filtered measurement of roll and pitch.
Note how the increment in thrust command $\widetilde{T}$ goes directly into the inner loop.

\subsection{Position control} \label{sec:poscontrol}
\begin{figure*}
	\centering
	\begin{tikzpicture}[auto, node distance=1cm, >=triangle 45]
		\small
		\draw
		node [input, name=input1] {}
		node [sum, right=0.7cm of input1] (sumb1){\minusa}
		node [gain, right=0.7cm of sumb1] (kxi){$K_\xi$}
		node [sum, right=0.7cm of kxi] (sumb2){\minusa}
		node [gain, right=0.7cm of sumb2] (kxidot){$K_{\dot{\xi}}$}
		node [block, right=0.7cm of kxidot] (AC) {$\mathrm{Outer\ INDI}$}
		node [block, right=0.7cm of AC] (integrator1) {$\frac{T_s z}{z - 1}$}
		node [block, right=0.7cm of integrator1] (integrator2) {$\frac{T_s z}{z - 1}$}
		node [output, right=1.0cm of integrator2] (out) {};
		\draw[->](input1) -- node {$\bs{\xi}_{\mathrm{ref}}$}(sumb1);
		\draw[->](sumb1) -- node {} (kxi);
		\draw[->](kxi) -- node {$\dot{\bs{\xi}}_{\mathrm{ref}}$} (sumb2);
		\draw[->](sumb2) -- node {} (kxidot);
		\draw[->](kxidot) -- node {$\bs{\nu}_{\ddot{\xi}}$} (AC);
		\draw[->](AC) -- node {$\ddot{\bs{\xi}}$} (integrator1);
		\draw[->](integrator1) -- node {$\dot{\bs{\xi}}$} (integrator2);
		\draw[->]([xshift=0.3cm]integrator1.east) -- ++(0,-1cm) -| node{} (sumb2);
		\draw[->]([xshift=0.3cm]integrator2.east) -- ++(0,-1.4cm) -| node{} (sumb1);
		\draw[->](integrator2) -- node{$\bs{\xi}$} (out);
		\draw
		;
		\normalsize

	\end{tikzpicture}
	\caption{The PD controller for the position.}
	\label{fig:position_loop}
\end{figure*}
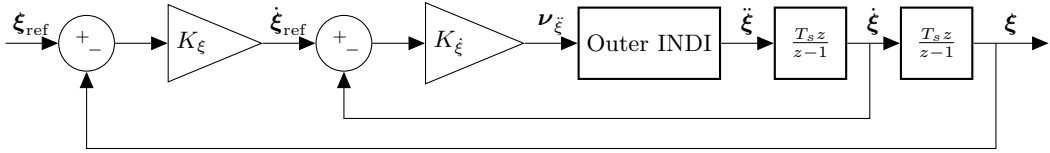

In the previous sections, it was shown how the linear acceleration can be controlled using an INDI controller.
To control the position of the MAV, an acceleration reference needs to be passed to the outer INDI controller that will steer the MAV towards its target position.
This can be done by a Proportional Derivative (PD) controller, as is shown in Figure \ref{fig:position_loop}.
The gains of this PD controller are manually tuned to give a fast response with little overshoot.

They depend mainly on two things: the update rate of the position estimate and the speed of the inner loop controller, which is only dependent on the actuator dynamics.
This is the case because all other components are inverted in the inversion step of the inner and outer loop.
Therefore, if these parameters are known in advance, one can come up with an estimate of the PD gains, for instance based on a pole/zero analysis.

\subsection{Adaptive control effectiveness}
In our previous work a Least Mean Squares algorithm was used to adapt the control effectiveness matrix of the rotors online \citep{smeur}.
Now, with respect to our previous work, a row is added to the control effectiveness matrix that predicts a change in thrust based on the actuator inputs.
This row of the control effectiveness matrix can also be adapted online, together with the rest of the matrix.
The LMS algorithm then becomes:
\begin{equation}
	\begin{array}{l}
		\boldsymbol{G}(k) = \boldsymbol{G}(k-1) \\  - \bs{\mu}_2 \left(\boldsymbol{G}(k-1)
			\left[ \begin{array}{c}\Delta \boldsymbol{\omega}_f \\
					\Delta \dot{\boldsymbol{\omega}}_f
			\end{array} \right]
	- \left[ \begin{array}{c} \Delta \dot{\bs{\Omega}}_f \\ \Delta T \end{array} \right] \right)
		\left[ \begin{array}{c}\Delta \boldsymbol{\omega}_f \\
				\Delta \dot{\boldsymbol{\omega}}_f
		\end{array} \right]^T
		\bs{\mu}_1
	\end{array}
	\label{eq:lms}
\end{equation}
where
\begin{equation}
\bs{G}(k) = \left[ \begin{array}{cc} \bs{G}_1(k) & \bs{G}_2(k) \end{array} \right]
	\label{eq:gdef}
\end{equation}

This means that the effectiveness of the motors with respect to the thrust can also be adapted online.
This can be important if the weight of the vehicle changes during flight, for instance when dropping a payload.
Given a flight with enough excitation of the control input and limited disturbances, the control effectiveness converges to the control effectiveness calculated offline.

\begin{table}
	\caption{Adaptation of the control effectiveness in the thrust axis in 10$^{-3}$ m/s$^2$/rpm}
	\centering
	\begin{tabular}{ccccc}
& M$_1$& M$_2$& M$_3$& M$_4$ \\
\hline
		Adaptation from -0.35  & -0.77 & -0.73 & -0.58 & -0.63\\
		Standard deviation & 0.025 & 0.024 & 0.012 & 0.016\\
		\hline
		Adaptation from -1.13  & -0.76 &   -0.73 &   -0.59 & -0.65\\
		Standard deviation & 0.021 & 0.033 & 0.030 & 0.027\\
		\hline
		Offline &  -0.76 & -0.72 & -0.57 & -0.63\\
		\label{tbl:adaptive}
	\end{tabular}
\end{table}

To demonstrate this, 10 flights were performed with the adaptive estimation enabled.
The flights were manually piloted, with constant maneuvering to ensure enough excitation.
Five of these started with the effectiveness of each rotor on the thrust set equal to -0.35, and five started with the effectiveness equal to -1.13.
From these flights, the adapted control effectiveness was recorded 30 seconds after takeoff.
Finally, one flight of two minutes was performed, and the logged flight data was processed offline with the method of Section \ref{sec:ctrl_eff} as a comparison.

The control effectiveness values are shown in Table \ref{tbl:adaptive}, averaged for the cases with five flights.
Clearly, after 30 seconds the control effectiveness of each motor on average has converged very close to the value calculated offline.
Additionally, it can be observed that the identified control effectiveness values differ quite a bit between motors.
These differences are naturally observed from the test-flight driven identification, whereas otherwise these details would be difficult to obtain.

In the next section, Figure \ref{fig:thrustcurve} shows the average thrust curve of the actuators.
One possible explanation of the differences observed in Table \ref{tbl:adaptive}, is that the motors are operating at different RPM.
However, the average RPM of the two minute flight was 126, 126, 125 and 123 for motors one through four respectively.

\subsection{Estimation of the thrust}
Throughout the derivation of the outer loop INDI controller, use is made of the thrust $T$, for instance in the matrix $G(\bs{\eta},T)$.
One possibility would be to measure the specific force in the body Z axis with the accelerometer, and use this as an estimate for the specific thrust ($\frac{T}{m}$).
This approach works well while hovering, but can lead to errors when there are other (aerodynamic) forces in the body Z axis.
These forces occur for instance at high speed steady flight, when the drone has a high bank angle.

\begin{figure}[htb]
	\centering
	\includegraphics[width=\columnwidth]{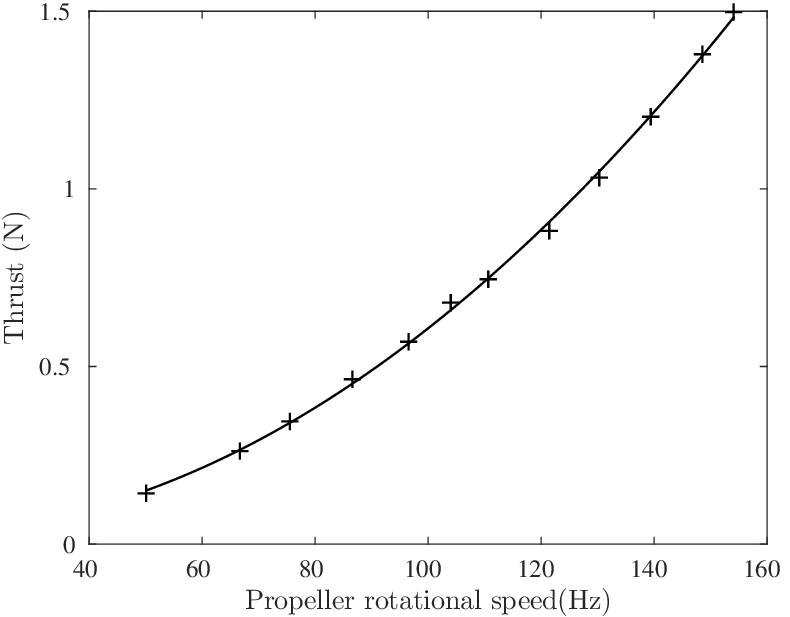}
	\caption{Thrust measurements for different rotor speeds along with a second order approximation.}
	\label{fig:thrustcurve}
\end{figure}

Therefore, static thrust measurements are used to model the thrust/rotational rate curve of the propellers.
The quadrotor was mounted upside down on a scale to obtain a measurement of the produced thrust.
The rotational rate of the propellers was obtained from the internal rpm measurement.
The resulting average thrust measurement per propeller as function of the rotational rate is shown in Figure \ref{fig:thrustcurve}.
A quadratic function showed a good fit with the data.
This function is used for the thrust estimate in the calculations of the controller.

Of course, the actual thrust of the propellers may be slightly different in a real flight, due to a different inflow.
Furthermore, since the propellers have a quadratic thrust curve, their control effectiveness changes depending on their current rotational rate.
In this paper, it is assumed that the control effectiveness of the rotors with respect to the specific force can be approximated by a static one.
This removes the need to recalculate $(\bs{G}_1+\bs{G}_2)^{-1}$ after every time step, enhancing the speed of the algorithm.

Errors introduced by these simplifications are expected to have low impact, because of the incremental nature of the controller.
If an increment in thrust does not lead to the desired acceleration, another increment is applied.
This way, small errors in the control effectiveness are handled naturally.

\subsection{Linearization} \label{subsec:nonlinearity}

In the previous section, a first order Taylor expansion was used to derive the INDI control law.
However, from Equation \ref{eq:tn} it can be seen that the force is actually very nonlinear in terms of roll and pitch.
In Equation \ref{eq:gtaylor} it can be seen that some of the derivatives can even change sign, for instance $\pd{\ddot{z}}{\phi}$ for different values of $\phi$.

What this means in practice is that if the increments in the input are large, because suddenly a large lateral acceleration is required, they will result in a different acceleration than intended.
This will be measured by the accelerometer, and subsequent increments in the inputs will eventually lead to the right acceleration.
But it might be more effective to implement a nonlinear method of finding increments in the input that give exactly the desired increment in the acceleration.

Since the nonlinear function of the inputs is available, it is possible to do a nonlinear inversion.
At the same time, it is preferable to keep the incremental structure, because there is no accurate estimate of the aerodynamic forces $\bs{F}$.
Referring to Eq. \ref{eq:newton}, it is possible to subtract the same formula a short time instant earlier:

\begin{equation}
	\begin{array}{ll}
		\ddot{\bs{\xi}} - \ddot{\bs{\xi}}_0  =& \bs{g} - \bs{g}_0 + \frac{1}{m}\bs{F}(\dot{\bs{\xi}},\bs{w}) - \frac{1}{m}\bs{F}(\dot{\bs{\xi}}_0,\bs{w}_0) \\
		&+ \frac{1}{m}\bs{T}_{N}(\bs{\eta},T) - \frac{1}{m}\bs{T}_{N}(\bs{\eta}_0,T_0)
	\end{array}
	\label{eq:newtondiff}
\end{equation}

It is assumed that changes in gravity and the aerodynamic forces are small during this small time instant:

\begin{equation}
	\ddot{\bs{\xi}} - \ddot{\bs{\xi}}_0  = \frac{1}{m}\bs{T}_{N}(\bs{\eta},T) - \frac{1}{m}\bs{T}_{N}(\bs{\eta}_0,T_0)
	\label{eq:newtondiffsimple}
\end{equation}

This equation relates an increment in thrust vector to an increment in acceleration.
The current thrust vector can be calculated based on the attitude and rotational rate of the rotors.
This gives an expression for the new thrust vector:

\begin{equation}
	\bs{T}_{N}(\bs{\eta},T) = m(\ddot{\bs{\xi}} - \ddot{\bs{\xi}}_0) + \bs{T}_{N}(\bs{\eta}_0,T_0)
	\label{eq:newtondiffinv}
\end{equation}

How the thrust vector depends on the thrust and attitude is described by Eq. \ref{eq:tn}.
A nonlinear inversion of this equation provides expressions for the thrust, roll and pitch commands:

\begin{equation}
	T = ||\bs{T}_N||
	\label{eq:tthrust}
\end{equation}

\begin{equation}
	\phi_c = \arcsin\left( \frac{\sin(\psi)T_{N_x}-\cos(\psi)T_{N_y}}{T}\right)
	\label{eq:tphi}
\end{equation}

\begin{equation}
	\theta_c = \arcsin\left( \frac{\cos(\psi)T_{N_x} + \sin(\psi)T_{N_y}}{T\cos{\phi}_c}\right)
	\label{eq:ttheta}
\end{equation}

This allows us to find a new attitude and thrust that will satisfy a desired acceleration, without linearizing the input function.
At the same time, the incremental structure is retained, as the new thrust vector is calculated based on the previous one.
For the nonlinear case, the same argument holds regarding the filtering as for the linearized case: if the acceleration is filtered, the other signals with subscript 0 should also be filtered.
This is shown below:

\begin{equation}
	\bs{T}_{N}(\bs{\eta},T) = m(\ddot{\bs{\xi}} - \ddot{\bs{\xi}}_f) + \bs{T}_{N}(\bs{\eta}_f,T_f)
	\label{eq:ttheta_filtered}
\end{equation}

\subsection{Filtering}
Both the measured accelerations as well as the rates are filtered to remove noise.
In the derivation of the INDI controller, it was shown that these signals should be filtered with the same filter.
This way, the delay in both loops is synchronized and the thrust increment can go from the outer loop to the inner loop.

Previously, it was shown that the filter choice has an effect on the disturbance rejection \citep{smeur}.
For the attitude loop, the response to a disturbance is given by $(1 - A(z)H(z)z^{-1})$.
By taking a filter with a higher cutoff frequency, and therefore less delay, disturbances will be rejected faster.
On the other hand, more noise will end up in the control signals.
Since the inner and outer INDI loops are connected and need to use the same filter, this trade-off should be considered for both loops simultaneously.
For the experiment, a filter is chosen with a $\omega_n =$ 50 rad/s and $\zeta =$ 0.55.

\subsection{Accelerometer bias} \label{subsec:bias}
The outer loop INDI controller is somewhat sensitive to accelerometer biases.
Because the accelerometer measurement is fed back to control the acceleration, an offset in the measurement will result in an offset in the actual acceleration as well.
This means that the quadrotor will maintain its position not at a position with zero error, but at a position where the position error times the P gain gives a required acceleration equal to minus the acceleration offset.

This problem does not arise in the inner loop, where the angular acceleration is bias free.
This is because in calculating the derivative of the rates, the bias disappears from the signal.
This is not the case for the outer loop, so it is necessary to estimate the accelerometer bias in order to remove it.
As a second measurement, the velocity and position from GPS or an indoor positioning system can be used.
The velocity measurement can be derived to obtain a bias free acceleration measurement.
Because of the low update rate, this signal is not really viable for feedback.

However, the acceleration obtained from velocity can be used to determine the accelerometer bias.
The derived acceleration has to be rotated to the IMU axes in order to take the difference with the acceleration measured by the IMU.
The accelerometer bias is a signal that is assumed to vary only very slowly.
This is why the acceleration difference is filtered with a second order filter with a natural frequency of 0.25 rad/s.
This removes all noise while keeping the important bias information.
This is only a very simple method of finding the accelerometer bias, it could alternatively be done with a Kalman filter.

\section{Windtunnel experiment} \label{sec:setup}
The goal of the experiment is to test how well the INDI controller can handle gust disturbances.
The experiment is performed indoors, such that there is a controlled environment in which repeatable experiments can be performed.
The source of the wind disturbance is the Open Jet Facility of the TU Delft Aerospace department.
It has a 2.85 m by 2.85 m cross section, and is capable of velocities up to 30 m/s.
A picture of the drone flying in front of the windtunnel is shown in Figure \ref{fig:experimentfoto}.
Because the windtunnel is not capable of rapidly increasing or decreasing its velocity, the quadrotor has to fly in and out of the windtunnel exhaust to simulate a gust.
This is done by letting the quadrotor alternate between two waypoints every 14 seconds, one being in the center of the windtunnel exhaust, and one being outside the windtunnel flow, at two meters distance.
For the experiment, the windtunnel is set to 10 m/s, while the maximum speed of the Bebop is 13 m/s, according to the manufacturer.
Needless to say, flying in and out of this flow is a simulation of a very strong gust.

Though this is the first time such an experiment is ever performed in a windtunnel, it bears some resemblance to an experiment performed earlier, using a big fan \citep{smeurgust}.
However, the maximum airspeed measured 1 m in front of the fan was only 4.0 m/s, a relatively small disturbance.
Moreover, the flow was not very homogeneous, as the airspeed was only 1.3 m/s at the center of the fan.
The use of a professional windtunnel makes the results much more quantitative.

The INDI loop basically controls the acceleration of the MAV, with a PD controller providing the acceleration reference.
From this perspective, it is a variation of a PID controller, where the INDI loop replaces the integral of the PID controller.
This is why the performance of INDI will be compared with a regular PID controller.
The PID controller is manually tuned to give the fastest response possible to a 5m step input, without oscillation.
Energy efficiency is not considered in the tuning of the controller.
Both the INDI as well as the PID outer loop controllers make use of the inner loop INDI controller for attitude control.
For the PID controller, the P, I and D gains work directly on the position and velocity to produce a reference roll, pitch and thrust, as is shown in Figure \ref{fig:pidctrl}.
Here $\bs{R}$ is a matrix defined by:
\begin{equation}
	\bs{R} = 
	\left[
	\begin{array}{cc}
		-\sin{\psi} & \cos{\psi} \\
		-\cos{\psi} & -\sin{\psi}\\
	\end{array}
\right]
	\label{eq:rotR}
\end{equation}

In tuning the PID gains, there is a trade-off to be made.
By increasing the integral gain, faster offset compensation can be obtained.
This way the quadrotor can adjust to the disturbance of the windtunnel faster.
However, with a high integral gain, the quadrotor will also have more overshoot in reference tracking tasks such as sudden position changes.
This trade-off is non-existent for the INDI controller.
Table \ref{tbl:params} and \ref{tbl:paramspid} present the parameters that have been used for the INDI and PID controllers respectively.

\begin{figure*}
	\centering
	\begin{tikzpicture}[auto, node distance=1cm, >=triangle 45]
		\small
		\draw
		node [input, name=input1] {}
		node [sum, right=0.7cm of input1] (sumb1){\minusa}
		node [gain, right=0.7cm of sumb1] (kxi){P}
		node [sum, right=0.7cm of kxi] (sumb2){\minusa}
		node [gain, right=0.7cm of sumb2] (kxidot){D}
		node [gain, below=0.5cm of kxidot] (igain){I}
		node [sum, right=1.2cm of kxidot] (sum){\suma}
		node [block, right=0.4cm of igain] (integrator) {$\frac{z}{z-1}$}
		node [block, right=0.7cm of sum] (R) {$\bs{R}$}
		node [block, right=1.1cm of R] (AC) {$\mathrm{Attitude\ control}$}
		node [output, right=1.0cm of AC] (out) {};
		\draw[->](input1) -- node {$\bs{\xi}_{\mathrm{ref}}$}(sumb1);
		\draw[->](sumb1) -- node {} (kxi);
		\draw[->](kxi) -- node {$\dot{\bs{\xi}}_{\mathrm{ref}}$} (sumb2);
		\draw[->](sumb2) -- node {} (kxidot);
		\draw[->]([xshift=0.2cm]sumb2.east) |- node {} (igain);
		\draw[->](kxidot) -- node {} (sum);
		\draw[->](igain) -- node {} (integrator);
		\draw[->](integrator) -| node {} (sum);
    \draw[->](sum) -- node {} (R);
    \draw[->](R) -- node {$\left[ \begin{array}{c} \phi_c \\ \theta_c \end{array} \right]$ } (AC);
		\draw[->]([yshift=-0.9cm]sumb1.south) -- node {$\bs{\xi}$} (sumb1);
		\draw[->]([yshift=-0.9cm]sumb2.south) -- node {$\dot{\bs{\xi}}$} (sumb2);
		\draw
		;
		\normalsize

	\end{tikzpicture}
	\caption{The horizontal PID controller used for comparison.}
	\label{fig:pidctrl}
\end{figure*}

\begin{table}[h]
	\centering
	\caption{INDI parameters.}
	\begin{tabular}{ccc}
		Parameter & Value & Unit\\
		\hline
		$\omega_n$ & 50 & rad/s\\
		$\zeta$ & 0.55 & \\
		$K_\Omega$ & 10.7 & (rad/s$^2$)/(rad/s)\\
		$K_\eta$ & 28.0 & \\
		$K_\xi$ & 0.7 & (m/s)/m\\
		$K_{\dot{\xi}}$ & 1.5 & (m/s$^2$)/(m/s)\\
		\label{tbl:params}
	\end{tabular}
\end{table}

\begin{table}[h]
	\centering
	\caption{PID parameters.}
	\begin{tabular}{ccc}
		Parameter & Value & Unit\\
		\hline
		P & 0.65 & (m/s)/m\\
		I & 0.11 & rad/(m/s)/s\\
		D & 0.2 & rad/(m/s)\\
		\label{tbl:paramspid}
	\end{tabular}
\end{table}

It is possible to do a crude comparison between the $K_\xi$ and $K_{\dot{\xi}}$ gains of the INDI controller and the P and D gains of the PID controller.
Around hover, the virtual control is related to the change in commanded attitude angle through a division by gravity, assuming small angles.
This means that $K_{\dot{\xi}}$ should be divided by 9.81, so $K_\xi$ and $K_{\dot{\xi}}$ become 0.70 (m/s)/m and 0.15 rad/(m/s) respectively.
For the PID controller, the corresponding P and D gains are 0.65 (m/s)/m and 0.20 rad/(m/s) respectively.
Though these gains are not exactly the same, the goal of this crude comparison is to show that both controllers have roughly the same gains.
Since the disturbance rejection properties will be considered, the integral gain will play the biggest role.

The MAV used for the experiments is the Bebop quadrotor from Parrot.
Instead of the stock firmware, it is running the Paparazzi open source autopilot system.
An infrared motion tracking system called 'Optitrack' was used to obtain position information.
This system can measure the drone's position with millimeter accuracy at a frequency up to 120 Hz.
But because the experiment should be realistic for outside scenarios and since most Global Positioning System (GPS) modules can only provide position updates at 4 Hz, the data was only sent to the drone at a frequency of 4 Hz.
The control algorithm, as well as the onboard accelerometer and gyroscope, were running at 512 Hz.
In an outdoor scenario, millimeter accuracy might not be achievable with off the shelve GPS modules.
But even though the position might be off in such a case, gusts will still be rejected the same way as in this indoor experiment, as the INDI controller is based on the accelerometer.


\subsection{Results} \label{sec:results}

\begin{figure}[htb]
	\centering
	\includegraphics[width=\columnwidth]{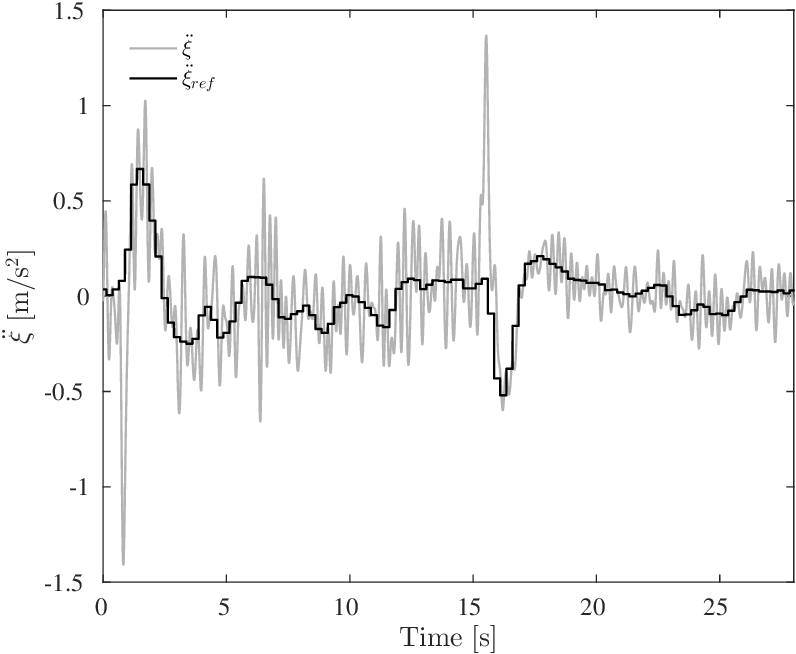}
	\caption{Acceleration in the north direction for the INDI controller.}
	\label{fig:accelx}
\end{figure}

First, consider Figure \ref{fig:accelx}.
It shows the acceleration in the north axis, which is the axis in which the windtunnel is blowing.
The acceleration is filtered with a second order filter with $\omega_n = 20$ rad/s and $\zeta = 0.7$.
The quadrotor starts besides the windtunnel, and at 0.0 seconds, the quadrotor is commanded to fly to the waypoint in front of the windtunnel.
The moment the quadrotor flies into the wind stream is clearly visible in the figure due to the large acceleration spike, deviating from the reference acceleration.
Due to this acceleration error, the INDI controller will increment the control inputs in order to make the acceleration track the reference again.
About half a second after the start of the disturbance, the acceleration coincides with the reference acceleration, effectively having counteracted the disturbance.
At that point, the quadcopter has built up a speed and position error in the north axis.
The quadrotor needs a positive acceleration after the disturbance to bring these errors back to zero.

The same thing happens at 15.4 seconds, when the quadrotor is commanded to fly out of the wind again.
Now the sudden absence of wind results in a disturbance in the opposite direction.
What also can be observed from this figure is that the accelerometer measures a more high frequency signal when flying in the wind.
This could be due to the airflow containing some turbulence.

\begin{figure}[htb]
	\centering
	\includegraphics[width=\columnwidth, trim=0 0cm 0cm 0cm]{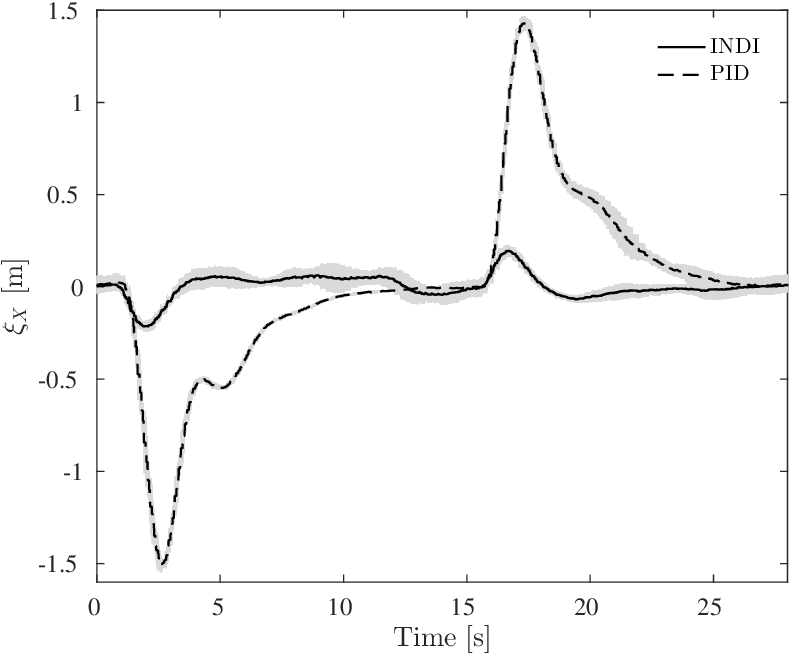}
	\caption{Position in the North direction for the INDI and PID experiment. The lines are the average of seven repetitions, and the shaded areas indicate one standard deviation.}
	\label{fig:posx}
\end{figure}

Figure \ref{fig:posx} shows the position along the $X_N$ axis, for both the INDI and PID controllers.
The figure shows the average of seven times the same maneuver, along with one standard deviation.
For INDI, it can be observed that a position error of 0.21 m occurs upon entering (2.0 seconds) and 0.20 m upon leaving (16.6 seconds) the windtunnel.
This position error is counteracted within three seconds after it occurred.

Compare this with the position for the PID controller in the same figure.
The maximum error is 1.51 m, and it takes longer for the position error to be counteracted as compared to the INDI controller.
One thing to note is that, when the vehicle is flying in front of the windtunnel and there are no changing disturbances, the PID controller shows less variance between flights than the INDI controller.
This difference may be attributed to the fact that the INDI controller is using the accelerometer for feedback.
Though the accelerometer measurement is filtered, it is still a bit noisy.
A filter with more high frequency attenuation could have been used, but this would make the disturbance rejection of the controller slightly slower, because such a filter has more delay.

\begin{figure}[htb]
	\centering
	\includegraphics[width=\columnwidth]{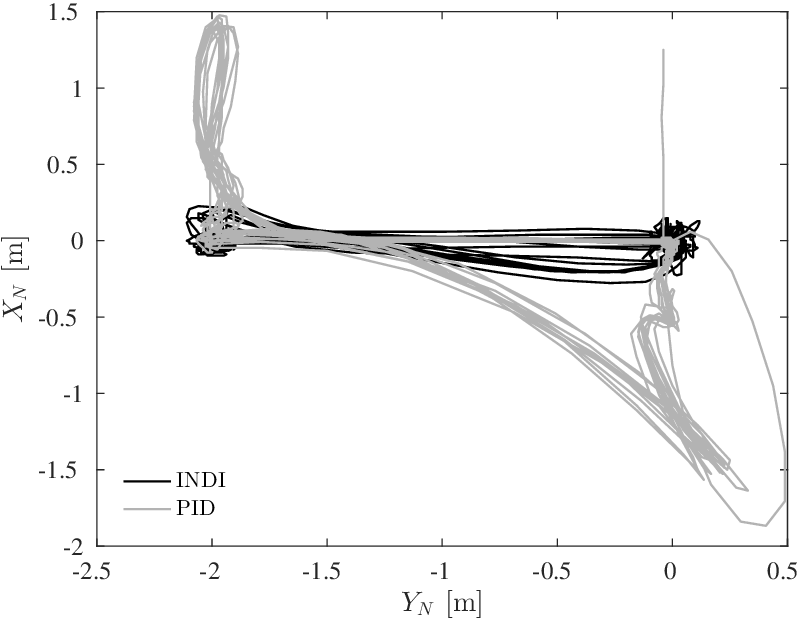}
	\caption{Top view of the experiment for the PID and INDI controllers. The windtunnel is blowing in the negative $X_N$ direction, and is located at $-1.425 < Y_N < 1.425$.}
	\label{fig:topview}
\end{figure}

A top view of the experiment is shown in Figure \ref{fig:topview}.
From this figure the difference in performance becomes apparent.
The figure shows the entire flight, from takeoff until landing.
For the PID controller, one can see how it is blown in the negative $X_N$ direction upon takeoff, entering the windtunnel, and how it overshoots in a straight line upon landing, leaving the windtunnel flow.
The INDI controller is able to cope much better with the sudden wind changes during taking off and landing.

\section{Outdoor takeoff with wind}

The experiment in the windtunnel is great from a scientific point of view, as it allows us to compare different controllers subject to exactly the same disturbance.
On the other hand, since an Optitrack system was used for position estimation, it might not be clear if the controller can provide the same performance in an outdoor scenario.
That is why a second experiment is performed; outdoors and with a standard off-the-shelve GPS receiver.

One of the situations in which an MAV needs to cope with a sudden wind disturbance, is during takeoff on a windy day.
When on the ground, the ground is compensating the drag from the wind.
But when the drone takes off, the wind force will accelerate the drone.
Therefore, a control action is needed to counteract the wind and maintain position.

Since the acceleration is measured with the accelerometer, it is expected that the INDI controller will compensate for the wind force very fast.
A PID controller that does not use this information, on the other hand, is expected to drift a bit, until it has gained some error in position and velocity that causes it to steer back.
The integrator part will remove the steady state error over time.

Like before, the first version of the Bebop quadrotor is used for this experiment.
As opposed to the second version of the Bebop, the first version that is used for this experiment has a low quality GPS.
With the built-in GPS, the disturbance rejection performance is hard to evaluate, as the position estimate will move around quite a bit, regardless if the drone is moving or not.
This is why the quadrotor is equipped with an external Ublox Neo M8N through a USB connection.
This GPS module is commercially available, and the second version of the Bebop even ships with this module built in.

Like with the windtunnel experiment, the case with outer loop INDI will be compared with a PID outer loop controller.
The PID controller has the same gains as in the windtunnel experiment, just like the P and D gains that produce the acceleration reference for the INDI controller are the same.

\subsection{Results}

On the day of the outdoor takeoff experiment an average wind speed of 5.1 m/s was reported by the Dutch Meteorological Institute (KNMI).
Over the course of one and a half hour, first twelve flights were performed with the PID controller, and then thirteen with the INDI controller.
The flights were performed one after the other, without breaks.
It is assumed that on average, the wind during the INDI flights was the same as during the PID flights, even though a fluctuation of the wind speed between flights was observed.

One of the flights with the INDI controller was rejected, as from the data it became clear that the state estimation filter had not converged prior to takeoff.
The state estimation error leads to a bias in the NED acceleration, which in turn leads to a position offset, as discussed above.

\begin{figure}[!t]
	\centering
	\includegraphics[width=\columnwidth]{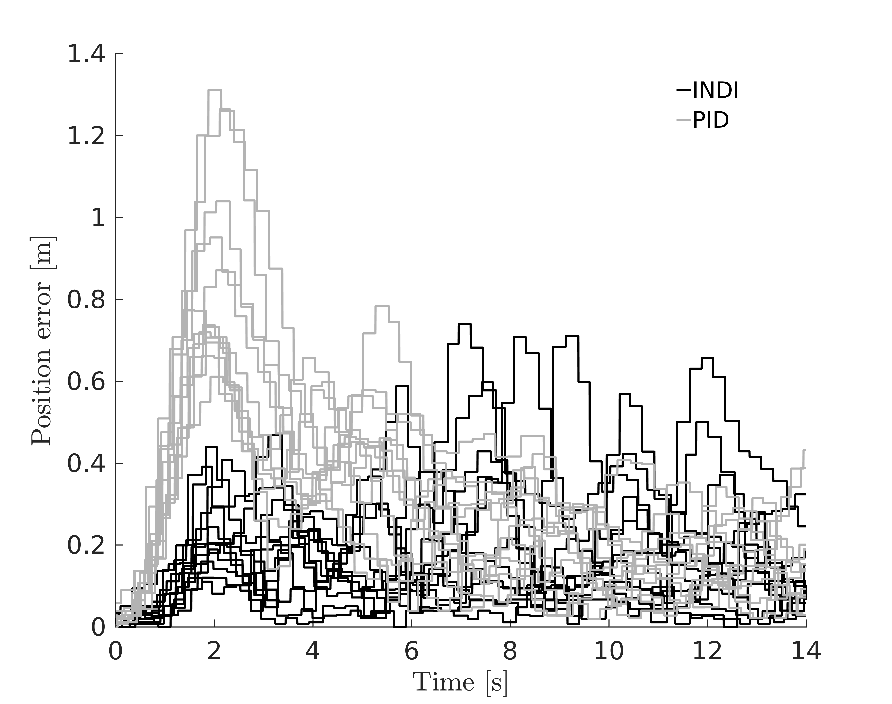}
	\caption{The horizontal position error during the outdoor takeoff experiment.}
	\label{fig:outdoorposerr}
\end{figure}

\begin{figure}[!t]
	\centering
	\includegraphics[width=\columnwidth]{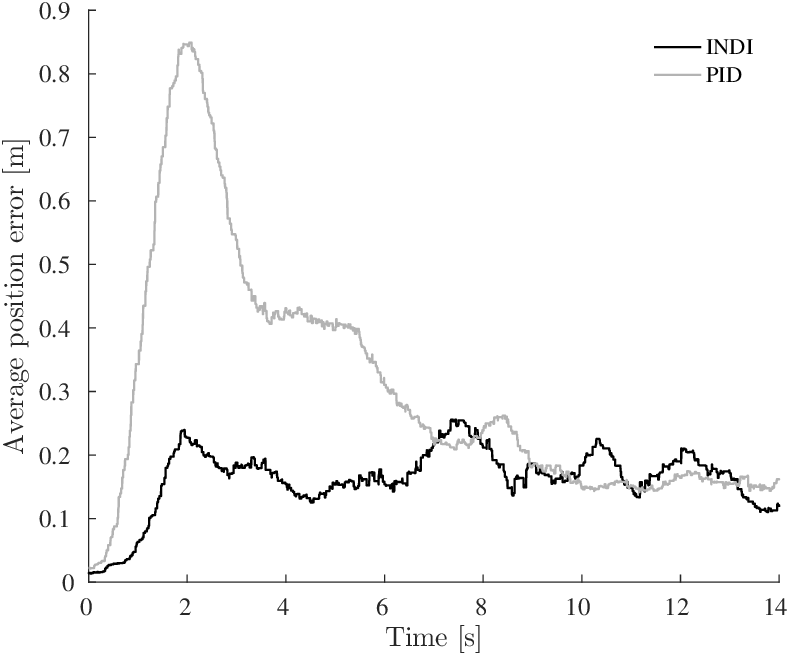}
	\caption{The average horizontal position error during the outdoor takeoff experiment.}
	\label{fig:avgoutdoorposerr}
\end{figure}

The position error can be seen in Figure \ref{fig:outdoorposerr}.
The average position error is shown in Figure \ref{fig:avgoutdoorposerr}.

The position reference was reset to the current position just before each flight, so all flights start with a position error close to zero.
As expected, during the takeoff INDI performs much better than the PID controller.
It can be seen from Figure \ref{fig:avgoutdoorposerr} that the INDI controller produces on average a maximum position error of 0.24 m as compared to 0.85 m for the PID controller.

Though Figure \ref{fig:avgoutdoorposerr} shows that the average error after some time is the same for both controllers, it appears from Figure \ref{fig:outdoorposerr} that there are some runs for the INDI controller with relatively large errors.
These errors are especially large if they are compared to the position error that is the result of the takeoff in the wind, which was expected to be the main disturbance.
Closer inspection of some of these datasets show that when these errors occur, the acceleration measured by the accelerometer does not correspond with the position and velocity measured by the GPS.
This may indicate that these errors are caused by GPS errors, perhaps upon changing between satellites.
This could perhaps be solved with a better state estimation algorithm, but that is beyond the scope of this research.

\section{Nonlinear increment}
As described in the section Implementation, the increment in thrust vector $\bs{T}_N(\bs{\eta},T)$ can also be computed without linearizing.
The linearization will give a small error if the virtual control $\bs{\nu}_{\ddot{\bs{\xi}}}$ is small.
But for large values of $\bs{\nu}_{\ddot{\bs{\xi}}}$, the error will be more significant.

It might be the case that, while such an incorrect increment in thrust vector is being executed and the quadrotor is rotating, a difference with the expected acceleration is already measured, and subsequent increments correct the thrust increment such that it will give the desired acceleration.
It will depend for a large part on the cutoff frequency of the measurement filter if this will be fast enough.
If the cutoff frequency of the filter is low, the delay may make the rejection of this disturbance too slow.

To assess whether the linearization is accurate enough for large acceleration changes, an experiment is devised.
In this experiment, from a hover initial condition, the quadrotor is commanded an acceleration of (0,4,0) m/s$^2$ for half a second, and then (0,-4,0) m/s$^2$ for another half second.
This way, it will go from accelerating in one direction, to accelerating in the other direction, resulting in a large acceleration change.
The maneuver takes place in the Y-axis, but considerable response differences in the Z-axis are expected.

This flight plan will be used for the Bebop quadrotor controlled with linearized INDI and nonlinear INDI.
For the linearized version the hypothesis is that the quadrotor will suddenly give very little thrust when the large change in acceleration is commanded, because of the derivative of the vertical component of $\bs{T}_N(\bs{\eta},T)$ with respect to thrust.
Due to this sudden decrease in thrust, the quadrotor is expected to slightly descend, before the vertical acceleration is measured and the thrust is increased again.
For the nonlinear version the hypothesis is that the thrust command will remain roughly the same, and little change in altitude is expected.

The experiment is performed in the indoor flight arena facility at the faculty of Aerospace Engineering in Delft.
The quadrotor is hovering based on the position feedback it receives from the tracking system.
It does not use this position information during the maneuvers, because the acceleration reference during these maneuvers is predefined.
The control effectiveness matrices $\bs{G}_1$ and $\bs{G}_2$ are determined prior to the experiment using the adaptive algorithm.
The experiment is repeated 25 times for both conditions.

\subsection{Results}
\begin{figure}[htb]
	\centering
	\includegraphics[width=\columnwidth]{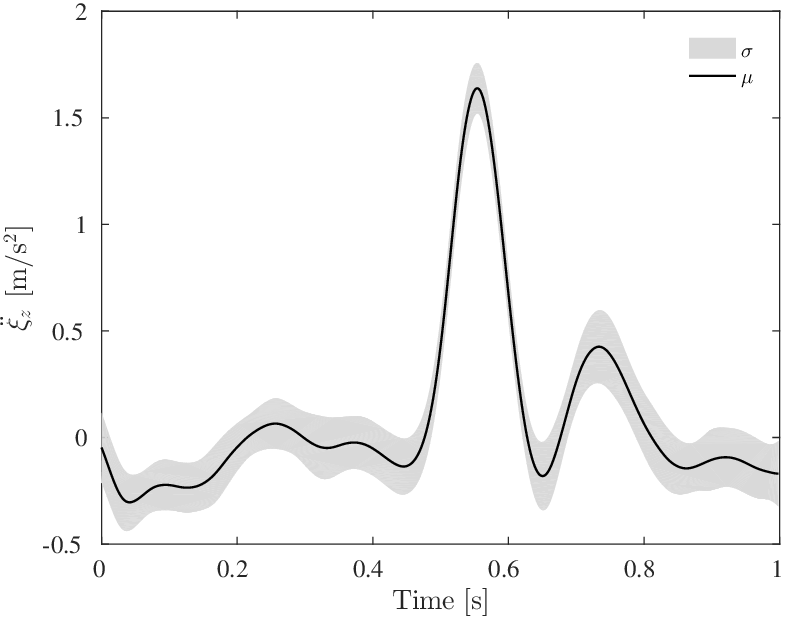}
	\caption{Acceleration in the NED Z axis with INDI increments calculated through linearization. From 25 experiments the mean $\mu$ is shown, along with one standard deviation $\sigma$.}
	\label{fig:lininc}
\end{figure}

Figure \ref{fig:lininc} shows the acceleration in the Z axis of the NED frame for the linearized case.
In the first two deciseconds, when the quadrotor is commanded to accelerate in the Y direction, the quadrotor has a slight upward acceleration, even though the thrust increment command at time zero is close to zero.
This can be explained with the fact that the inner loop control effectiveness inversion is linear, and will add as much RPM on one side of the quadrotor as on the other side to make it bank.
Actually, the relation between RPM and thrust is nonlinear (see Figure \ref{fig:thrustcurve}), and if all propellers are spinning equally fast, a bank command will therefore result in a slight thrust increase.

More profound is the downward acceleration that happens after half a second, when the quadrotor has to accelerate in the -Y direction.
Because the quadrotor is banking to facilitate the acceleration in the +Y direction, the derivative of the vertical acceleration with respect to the bank angle is negative (upward acceleration) for a reduction in bank angle.
Therefore, even though eventually around the same thrust is required, initially the thrust is reduced significantly, resulting in a downward acceleration.

\begin{figure}[htb]
	\centering
	\includegraphics[width=\columnwidth]{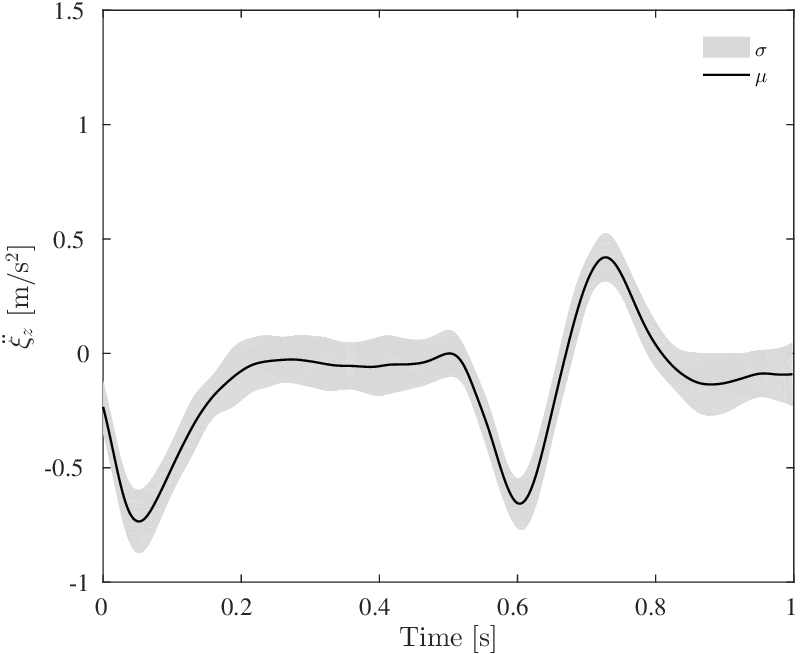}
	\caption{Acceleration in the NED Z axis with INDI increments calculated through nonlinear calculation. From 25 experiments the mean $\mu$ is shown, along with one standard deviation $\sigma$.}
	\label{fig:nonlininc}
\end{figure}

Compare this with Figure \ref{fig:nonlininc}, which shows the nonlinear case.
Here a larger acceleration is visible in the first two deciseconds.
This is caused by the fact that the actuator dynamics are faster than the rotational dynamics.
The nonlinear increment is calculated for the tilted thrust vector, therefore a positive thrust increment is commanded by the outer loop INDI controller.
However, the rotational dynamics are slower than the thrust dynamics.
Therefore, the thrust is increased already before the final attitude is attained.
This causes the vehicle to accelerate upwards initially.

After half a second, when the large acceleration change is commanded, the response is quite different from the linear case.
Instead of acceleration downward, the vehicle accelerates upward.
This can be explained by recognizing that the quadrotor will need more or less the same bank angle to accelerate with the same amount in the other direction.
This means that the same thrust is needed.
However, while the vehicle is rotating, it passes the point of zero bank angle, for which it actually needs less thrust to avoid a vertical acceleration.
This can explain that the vehicle accelerates upward, reduces thrust, and then overshoots to downward acceleration when it reaches the bank angle at which increased thrust is needed.

Comparing, the nonlinear implementation results in a vertical acceleration that averages better to the intended zero m/s$^2$.
However, there is still some unintended vertical acceleration present.
This is mainly attributed to the nonlinear way that input increments are realized, as is described above.
Additionally, in the inner loop the nonlinear thrust curve of Figure \ref{fig:thrustcurve} is not taken into account, which may lead to some error.

The acceleration changes in the experiment were the largest possible without introducing saturation in the actuators.
Of course, the larger the acceleration change, the larger the nonlinear effects.
To analyze if the difference is more significant for larger acceleration changes, more experiments are necessary.


\section{Conclusions}
In this paper, the control of a micro air vehicle using Incremental Nonlinear Dynamic Inversion (INDI) has been demonstrated for both the inner loop (attitude control) as well as the outer loop (position control) in a cascaded fashion.
The disturbance rejection performance of the resultant controller is examined by flying in and out of a 10 m/s windtunnel flow, showing a more than 7 times lower maximum position deviation than a comparable PID controller.
From other experiments it was concluded that the control method is applicable outdoors and that also the control effectiveness of the actuators on the thrust can be adapted online.
The nonlinear calculation of the thrust vector increment reduces the maximum error in vertical acceleration tracking for an aggressive maneuver, but further research is needed to establish if this method yields significant benefits.

The controller derived in this paper can provide two main benefits.
Firstly, the disturbance rejection properties can allow vehicles to operate close to obstacles in gusty environments.
Secondly, all parameters, except the position and velocity gains, can be determined based on a test flight and an identification of the actuator dynamics.
This makes implementation on new platforms easy and straightforward.
Finally, the online adaptation of the control effectiveness can account for changes made to an airframe, and offer even more ease of use.

\subsection{Future work}
The investigation of the effects of linearization in the outer loop indicates that the performance of the inner loop may be further improved by considering the nonlinear relation of rpm and thrust.
This could be done by linearly calculating increments for the inner loop, and then use the nonlinear mapping of Figure \ref{fig:thrustcurve} to map the linear increments to the correct nonlinear increments.

Though the inner and outer loop INDI controllers are quite robust, a situation that can still lead to instability is saturation of the actuators.
In this case, doing the control allocation through the inverse of the control effectiveness matrix and saturating the resultant control vector, leads to a suboptimal realization of the control objective, because some axes are more important than others.
Preliminary research shows that this problem can be solved by taking the axis priorities into account when calculating the control vector \citep{smeur2017}.

In the derivation of the outer loop INDI controller, it was assumed that changes in $\psi$ would be small, such that the derivative of the thrust vector with respect to $\psi$ could be neglected.
A better solution may be to switch to a different Euler angle rotation order.
Instead of the common ZYX order, which is used in this paper, a better choice may be the XYZ order.
This will remove any dependency of the thrust vector on the angle $\psi$.

Furthermore, this control method will be applied to hybrid UAVs, that combine vertical takeoff and landing with fast forward flight using a wing.
These vehicles are very prone to be disturbed due to their large aerodynamic surfaces, and INDI is especially good at disturbance rejection.
An INDI attitude controller has been used for a tilt rotor vehicle in simulation by \cite{francesco2016}, but they used a model instead of angular acceleration feedback, which means that the disturbance rejection properties are lost.
\cite{bronz2017} did preliminary experiments with a tailsitter based on the algorithms in this paper, showing promising results.

\section*{Funding}
This research was funded by the Delphi Consortium.

\section*{Acknowledgment}
The authors would like to thank Matej Karasek, for his help with the windtunnel experiment, and Freek van Tienen, for porting the Paparazzi software to the Bebop quadrotor.

\bibliography{bibliography.bib}

\end{document}